\documentclass[11pt,a4paper]{article}
\usepackage[final]{acl}
\usepackage{times}
\usepackage{latexsym}

\usepackage{amsmath}
\usepackage{amsfonts}
\usepackage[linesnumbered,ruled,vlined]{algorithm2e}
\usepackage{graphicx} 
\usepackage{booktabs}
\usepackage{array}
\usepackage{enumitem}
\usepackage{amsthm}
\usepackage{algpseudocode}

\usepackage[switch]{lineno}
\usepackage[utf8]{inputenc}

\usepackage{eqparbox}
\usepackage[nopar]{lipsum}
\usepackage{multirow}
\usepackage{makecell}

\usepackage{xcolor}

\usepackage{scrextend}
\deffootnote[.25in]{.25in}{.15in}{\makebox[.25in][r]{\thefootnotemark .\hspace{.15in}}}

\makeatletter

\newtheorem*{proof*}{Proof}

\usepackage{listings}
\usepackage{color}
\definecolor{codegreen}{rgb}{0.3,0.5,0.0}
\def\@fnsymbol#1{\ensuremath{\ifcase#1\or \dagger\or *\or \ddagger\or
   \mathsection\or \mathparagraph\or \|\or **\or \dagger\dagger
   \or \ddagger\ddagger \else\@ctrerr\fi}}

\title{A Mutually Reinforced Framework for Pretrained Sentence Embeddings}

\author{Junhan Yang$^1$\thanks{Work is done during the internship at Microsoft.}, ~Zheng Liu$^2$\thanks{Corresponding author.}, ~Shitao Xiao$^3$\footnotemark[1], ~Jianxun Lian$^2$, ~Lijun Wu$^2$, ~Defu Lian$^1$, ~Guangzhong Sun$^1$, ~Xing Xie$^2$\\
  1: University of Science and Technology of China, Hefei, China \\ 
  2: Microsoft Research Asia, Beijing, China \\
  3: Beijing University of Posts and Telecommunications, Beijing, China \\
  \texttt{yangjun2@mail.ustc.edu.cn} \\
  \texttt{\{zhengliu,jialia,lijuwu,xingx\}@microsoft.com} \\
  \texttt{stxiao@bupt.edu.cn} \\
  \texttt{\{liandefu,gzsun\}@ustc.edu.cn}}

\begin{document}
\maketitle
\begin{abstract}
The lack of labeled data is a major obstacle to learning high-quality sentence embeddings. Recently, self-supervised contrastive learning (SCL) is regarded as a promising way to address this problem. However, the existing works mainly rely on hand-crafted data annotation heuristics to generate positive training samples, which not only call for domain expertise and laborious tuning, but are also prone to the following unfavorable cases: 1) \textit{trivial} positives, 2) \textit{coarse-grained} positives, and 3) \textit{false} positives. As a result, the self-supervision's quality can be severely limited in reality. 

In this work, we propose a novel framework \textbf{InfoCSE}\footnote{\scriptsize InfoCSE: mutually re\textbf{Info}rced self-supervised \textbf{C}ontrastive learning of \textbf{S}entence \textbf{E}mbeddings.} to address the above problems. Instead of relying on annotation heuristics defined by humans, it leverages the sentence representation model itself and realizes the following iterative self-supervision process: on one hand, the improvement of sentence representation may contribute to the quality of data annotation; on the other hand, more effective data annotation helps to generate high-quality positive samples, which will further improve the current sentence representation model. In other words, the representation learning and data annotation become mutually reinforced, where a strong self-supervision effect can be derived. Extensive experiments are performed based on three benchmark datasets, where notable improvements can be achieved against the existing SCL-based methods. 

\end{abstract}

\section{Introduction}


Sentence embeddings are critical for information retrieval services, such as recommender systems and search engines. With the development of deep learning techniques, the DNN-based representation models have been widely applied to enhance the quality of sentence embeddings. However, deep neural networks usually require a great deal of labeled data, which is unrealistic for those cold-start scenarios. In recent years, Self-supervised Contrastive Learning (\textbf{SCL}) is recognized as a promising way of addressing the above problem, with which an immense amount of positive training samples can be introduced from the unlabeled data. Despite the achieved progress so far, the existing data annotation strategies are mainly based on hand-crafted heuristics. For one thing, the heuristic rules need to be designed with domain expertise and laborious tuning. For another thing, the heuristically annotated positive samples are still prone to the following undesirable cases.

\begin{figure}[t]
\centering
\includegraphics[width=0.98\linewidth]{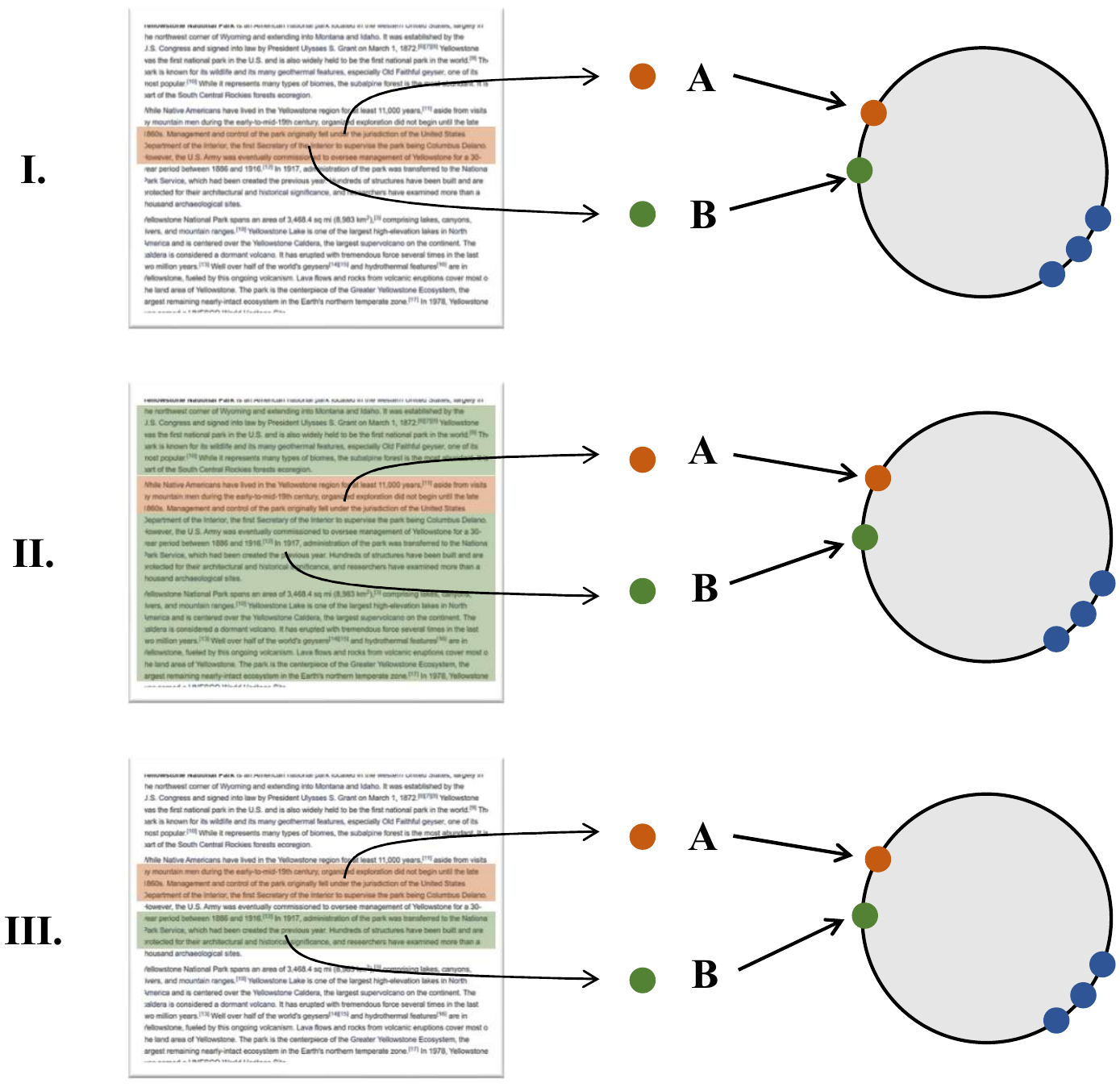}
\vspace{-5pt}
\caption{\small Typical methods to annotate positive samples. (I) different views of the same sentence; (II) one sentence and its context; (III) the neighboring sentences within the same document. The representation model is learned to project the positive samples (orange and green) close to each other and keep them discriminated from the negative samples (blue).}
\vspace{-18pt}
\label{fig:1}
\end{figure}

\textit{Trivial positives}. One typical class of methods is to generate different views for the same sentence (Figure \ref{fig:1} I.), where each pair of views are annotated as a positive sample. For example, the input sentence is cutoff and shuffled in \cite{yan2021consert} for different views; in \cite{liu2021fast}, a span of the sentence is randomly masked; and in \cite{gao2021simcse}, the sentence is encoded twice with different dropout masks. Although different views of the same sentence are semantically correlated, their positive relationships can be too easy to predict due to the high lexical similarity. As a result, it is unfavorable for the representation model to learn the in-depth semantics of the sentences. 

\textit{Coarse-grained positives}. Another class of methods leverages the correlation between a sentence and its context (Figure \ref{fig:1} II.). One representative strategy is the Inverse Cloze Task (ICT) \cite{guu2020realm,chang2020pre}, where one sentence and the rest of the document are treated as a positive sample. A relevant method is the Contrastive Predictive Coding (CPC) \cite{oord2018representation}, where one sentence and its preceding content are treated as a positive sample. However, such correlation relationships can be coarse-granular: considering that the document's semantics are usually diversified, the sentence may only be correlated with a limited part of its context. 

\textit{False positives}. People also label different sentences as positive samples based on certain rules; e.g., it is a common practice to treat neighboring sentences as positive samples \cite{kiros2015skip,hill2016learning,cer2018universal,iter2020pretraining} (a special case of the above class, as individual sentences are sampled from the context). Although such sentences enjoy certain external similarities, there is no guarantee of their semantic correlations. Thus, it is inevitable to generate false positives, which are harmful to self-supervision. 

To address the existing limitations, we propose a novel self-supervision framework \textbf{InfoCSE} to pretrain sentence embeddings. Unlike the existing methods which intensively rely on heuristic strategies, InfoCSE leverages its own representation model for data annotation. \textit{It initializes the representation model based on weakly annotated positive samples. Then, it uses the representation model to revise the annotation strategy. This will give rise to more quality positive samples, with which the representation model can be further improved.} On top of the above iterative process, the data annotation and sentence representation become mutually reinforced, which substantially enhances the self-supervision's effect.


The discovery of high-quality positive samples is the key for InfoCSE. The positive sample is expected to satisfy two properties: 1) the inputs should have strong \textit{semantic correlation} so as to ensure the correctness of supervision signal; 2) the inputs are desired to be \textit{lexically different}, which facilitates the generation of non-trivial positives. Driven by this demand, we design the \textbf{Intra-Document Clustering} (\textbf{IDC}) algorithm to exploit positive samples. Particularly, two distinct sentences from the same document are treated as a ``candidate positive sample'', knowing that they are different in lexicons but potentially related in semantics. The candidate positives are verified based on clustering: sentences in the same documents are clustered based on their embeddings' closeness (\textit{measured by the current sentence representation model}); then, two sentences in the common cluster are annotated as positive because their semantic correlation is relatively higher than the majority of sentences in the same document. A simple but effective graph partition-based method is designed for clustering, where similar sentences can be grouped together with very little cost.

The basic InfoCSE mainly targets zero-shot scenarios, where no labeled data is available. On top of this foundation, we extend InfoCSE with SimCLRv2 \cite{chen2020big} as \textbf{InfoCSE++}, which further improves the effectiveness in few-shot scenarios (a few labeled data is given). The major contributions are listed as follows.


\begin{itemize}[leftmargin=3pt,noitemsep,topsep=0pt]
    \item To the best of our knowledge, InfoCSE is the first work that brings the mutually reinforced data annotation and representation learning to self-supervised sentence embeddings. With InfoCSE, high-quality positive samples can be iteratively generated, which substantially enhances the self-supervision's performance.
    \item We design IDC-based annotation algorithm, with which lexically different but semantically relevant sentences can be effectively extracted from documents as positive samples.
    \item We extend our framework as InfoCSE++, which further improves the sentence embeddings' quality in few-shot scenarios.
    \item We evaluate our proposed methods with three benchmark datasets, where notable improvements can be achieved against the SOTA SCL-based sentence embeddings. 
\end{itemize}

\section{Related Work}
Sentence representation is a fundamental issue in NLP and IR communities. With the development of deep learning, DNN-based representation models have been widely used for the learning of sentence embeddings \cite{huang2013learning,hill2016learning,cer2018universal,reimers2019sentence}. However, the DNN-based approaches usually call for a great deal of training data, which poses a severe challenge for those cold-start scenarios with very few labeled data. To mitigate this problem, various pretraining approaches have been proposed. For example, sentence embeddings can be pretrained on top of transfer learning: the representation model is supervisedly learned on the source domains, and then applied to generate sentence embeddings for the target domains. The natural language inference (NLI) datasets \cite{bowman2015large,williams2017broad} were found to be appropriate source domains, whose pretrained sentence embeddings achieve quite promising performances on various downstream tasks \cite{hill2016learning,cer2018universal,reimers2019sentence}. Besides, people also take advantage of self-supervised learning, where tremendous amounts of unlabeled data can be leveraged. In \cite{hill2016learning,wang2021tsdae,lu2021less}, the sequential denoising auto-encoder has been adopted as a pretraining task for sentence representation; compared with the conventional pretraining tasks like MLM \cite{Devlin2019BERT,Liu2019Roberta}, the underlying semantics about the sentences can be better represented by the generated embeddings. 

One of the most emphasized classes of pretrained sentence embeddings is based on self-supervised contrastive learning (SCL) \cite{chen2020simple,chen2020big,grill2020bootstrap}. The underlying intuition about SCL is that the semantically correlated data instances can be generated from unlabeled data with pre-defined heuristics. By learning to discriminate a large amount of correlated and non-correlated data instances, the representation model will be able to capture the underlying semantics of the data with concise and expressive embeddings. In recent years, various SCL-based sentence embeddings have been proposed. In \cite{wu2020clear,zhang2021bootstrapped,yan2021consert,liu2021fast,gao2021simcse}, an input sentence is cast into different views, with operations like deletion, reordering, span masking, etc. Each pair of views may serve as a positive sample for self-supervised contrastive learning. In \cite{lee2019latent,chang2020pre,guu2020realm}, the sentences and their contexts within the documents are treated as positive samples. And in \cite{kiros2015skip,iter2020pretraining,di2021exploiting}, two neighboring sentences may also be regarded as a positive sample. As we have discussed, all these data annotation strategies are based on hand-crafted heuristic rules. For one thing, it requires laborious exploration and domain expertise. For another thing, it is prone to unfavorable annotation results, including trivial positives, coarse-grained positives, and false positives.
Our work is fundamentally different from the existing methods: instead of relying on heuristics defined by humans, it leverages the representation model itself for data annotation. By making the data annotation and sentence representation mutually reinforced, high-quality positive samples can be iteratively excavated to enhance self-supervision. 


\section{Methodology}

We leverage self-supervised contrastive learning to pretrain the sentence embeddings with unlabeled data. Without loss of generality, the learning objective is formulated as follows:
\begin{equation}\label{eq:1}
    \mathcal{L}(a,p) = \frac
    {\exp\{\langle f(a), f(p) \rangle\}}
    {\sum_{n{\neq}a}\exp\{\langle f(a), f(n) \rangle\}}.
\end{equation}
Here, $a$ (anchor) and $p$ (positive) denote the two inputs for the positive sample, and $n$ is a negative sample to $a$; $f(\cdot)$ is the representation model, which encodes the sentence into embedding, and $\langle\cdot\rangle$ is the inner-product operator. In this work, a collection of documents $\mathbf{D}$ is utilized for the pretrained sentence embeddings. For each document $d\in \mathbf{D}$, the positive training samples are generated from it for self-supervised contrastive learning. 

\begin{figure*}[t]
\centering
\includegraphics[width=0.92\textwidth]{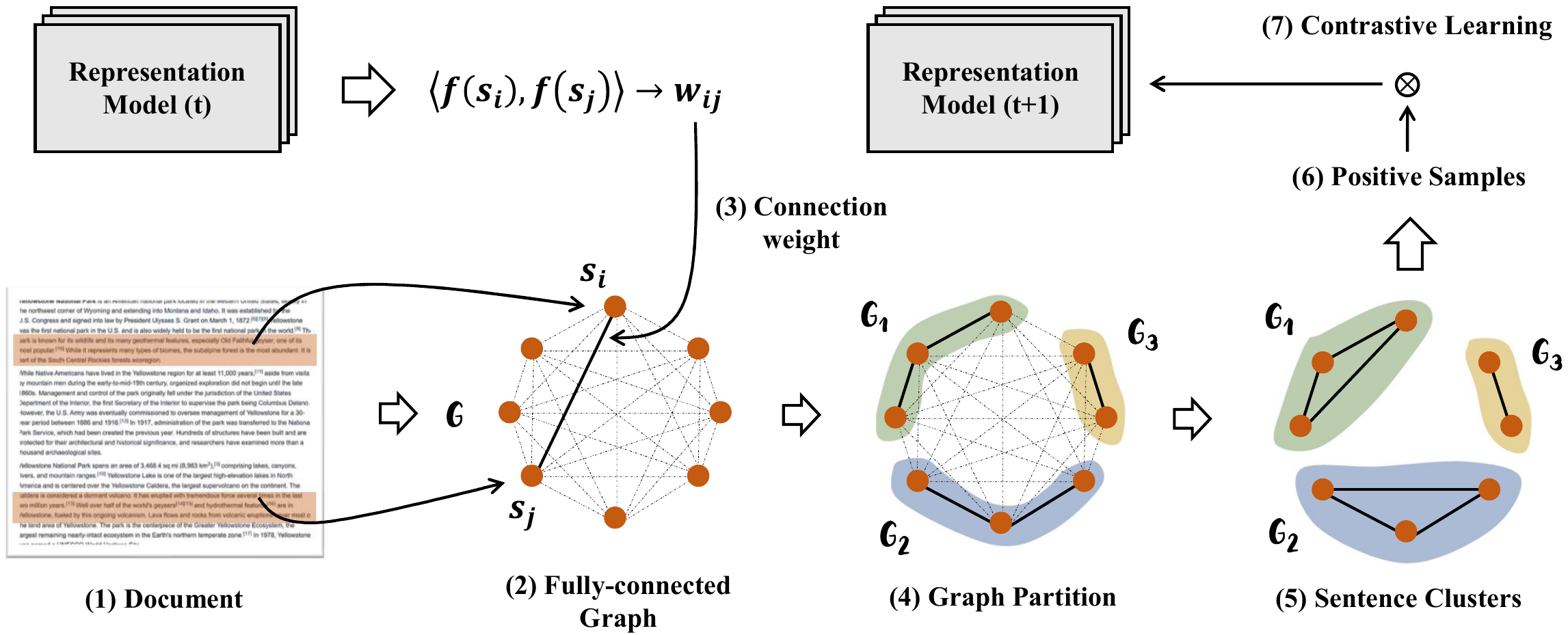}
\vspace{-5pt}
\caption{\small Overview of InfoCSE and IDC. (1-2): Sentences within the same document are organized as a fully-connected graph; (3-5): the connections between sentences are weighted by the current representation model (iteration t), based on which the graph is partitioned into clusters; (6-7): all pairs of sentences within the same clusters are annotated to be positive samples, which are used for the contrastive learning of the representation model for the next iteration t+1.}
\vspace{-15pt}
\label{fig:2}
\end{figure*}

\subsection{Intra-Document Clustering (IDC)}
We excavate \textit{semantically close} but \textit{lexically different} sentences from the documents, which may serve as reliable and non-trivial positive samples for self-supervised contrastive learning. To facilitate the effective discovery of such sentences, the Intra-Document Clustering is devised (Figure \ref{fig:2}).

Firstly, each pair of distinct sentences ($s_i$, $s_j$) within document $d$ is taken as a ``candidate positive sample'': for one thing, distinct sentences are lexically different; for another thing, the two sentences are potentially related in semantics given that they are from the same document. Secondly, the candidate is verified through clustering: sentences within the same document are grouped into clusters based on their correlation measured by the sentence representation model $f(\cdot)$ at present; if the sentences $s_i$ and $s_j$ are included within the same cluster, the candidate ($s_i$, $s_j$) is confirmed to be a positive sample.

$\bullet$ \textbf{Partition-based clustering}. The clustering operation needs to be flexible: each document may have a different degree of diversity about its content (the more diversified content, the more clusters can be formed); thus, it's inappropriate to pre-define the number of clusters, like $k$-Means. Besides, the clustering needs to be efficient so as to reduce the running cost. For these purposes, partition-based clustering is designed. 


We organize the sentences within the same document as a fully connected graph $\mathcal{G}$. For each pair of sentences $s_i$ and $s_j$, their connection is weighted by the representation model $f(\cdot)$:
\begin{equation}\label{eq:2}
    w_{ij} \leftarrow \langle f(s_i), f(s_j) \rangle,
\end{equation}
where $w_{ij}$ is the connection weight. Then, the connections will be pruned unless any of the following relationships is satisfied:
\begin{equation}\label{eq:3}
w_{ij} \in \text{top-K}\{w_{i*}\}, ~\text{or}~ w_{ij} \in \text{top-K}\{w_{*j}\}.
\end{equation}
That is to say, $w_{ij}$ will be preserved if $s_j$ is among the top-K relevant sentences to $s_i$, or $s_i$ is one of the top-K relevant sentences to $s_j$. The default value of K is set to $1$, which is sufficient to achieve competitive performances in experiments. After the pruning of connections, the original graph can be partitioned into $L$ clusters: $\{\mathcal{G}_l\}_L$, where each cluster $\mathcal{G}_l$ is a connected graph component. 

The partition-based clustering is experimentally verified to be \textit{effective}. Besides, it is highly \textit{efficient}: it just calls for one pass of scan for the graph nodes, whose time cost is almost ignorable. 

\subsection{InfoCSE}

The InfoCSE is performed as Figure \ref{fig:2}. The overall workflow can be divided into two parts: the generation of positive samples with the current representation model (\textit{t}): step (1--6) (``\textit{t}'' stands for the current iteration), and the contrastive learning for the next representation model (\textit{t}+1): step (7).

$\bullet$ \textbf{Positive samples}. The positive samples are generated from Intra-Document Clustering, 
where the current representation model (\textit{t}) is utilized. The sentences within each of the input documents are organized as a fully-connected graph $\mathcal{G}$. Each of the connections in $\mathcal{G}$ is weighted based on the similarity of the sentence embeddings, as Eq. \ref{eq:2}. Then, the connections are pruned based on the relationships in Eq. \ref{eq:3}, where the graph can be partitioned into clusters $\{\mathcal{G}_l\}_L$. Finally, each pair of sentences within the same cluster form a positive sample; all such sentence pairs become the positive samples from document $d$:
\begin{equation}\label{eq:4}
    \mathbf{P}_d: \{(s_i,s_j)|\exists l: s_i,s_j\in\mathcal{G}_l\}.
\end{equation}

$\bullet$ \textbf{Contrastive learning}. With the collection of positive samples from all the documents, the representation model for the next iteration (\textit{t}+1) is learned to minimize the contrastive loss in Eq. \ref{eq:1}:
\begin{equation}\label{eq:5}
    f^* \leftarrow argmin. \sum_{d\in\mathbf{D}}\sum_{(s_i,s_j)\in\mathbf{P}_d} l(s_i,s_j).
\end{equation}
Note that the positive sample is unordered: $(s_i,s_j)$ and $(s_j,s_i)$ are regarded as the same and used for once. As for negative samples: following the typical treatments \cite{karpukhin2020dense,chen2020simple}, we use in-batch negative sampling, where sentences (from different documents) within the same batch are utilized as the negative samples. 

$\bullet$ \textbf{Model initialization}. InfoCSE needs to get started from an initialized sentence representation model. This can be done by running an arbitrary SCL-based sentence representation algorithm. In this work, we leverage CONPONO \cite{iter2020pretraining}, where each sentence ($s_k$) is encoded to predict other sentences within the same context (i.e., $s_{k-i}$, $s_{k+j}$), in contrast to the random negatives. It's worth mentioning that InfoCSE is experimentally verified to be ``initialization robust'': other popular self-supervision methods, despite differences in initial performances, may also be utilized and will always converge to competitive performances after a few rounds of iterations. 

\setlength{\textfloatsep}{10pt}
\begin{algorithm}[t]
\caption{InfoCSE}\label{alg:1}
    \LinesNumbered 
    \SetKwInOut{KwIn}{Input}
    \SetKwInOut{KwOut}{Output}
    \KwIn{document set $\mathbf{D}$}
    \KwOut{representation model $f(\cdot)$}
    \Begin{
        $f \leftarrow$ model initialization\;
        \While{not converge}{
            \For{$d\in\mathbf{D}$}{
                get $\mathbf{P}_d$ based on $f$, as Eq. \ref{eq:4}\;
            }
            learn $f^*$ with $\{\mathbf{P}_d\}_{\mathbf{D}}$, as Eq. \ref{eq:5}\;
            replace $f$ with $f^*$\;
        }
        return $f$\;
    }
\end{algorithm}

$\bullet$ \textbf{Algorithm}. The workflow of InfoCSE is summarized as Alg \ref{alg:1}. Firstly, the representation model is initialized. Starting from the initialized model, the iterations are launched for the positive samples' generation and contrastive learning: the current model $f$ is used to generate positive samples from all the documents; then, the model for the next iteration $f^*$ is learned by minimizing the contrastive loss w.r.t. all the training samples. We leverage the validation set to monitor the training process: once the validation performance stops growing, the iteration will be terminated and the optimal representation model is returned. We experimentally find that InfoCSE is quick to converge, usually within 3 iterations. 

\begin{figure}[t]
\centering
\includegraphics[width=1.0\linewidth]{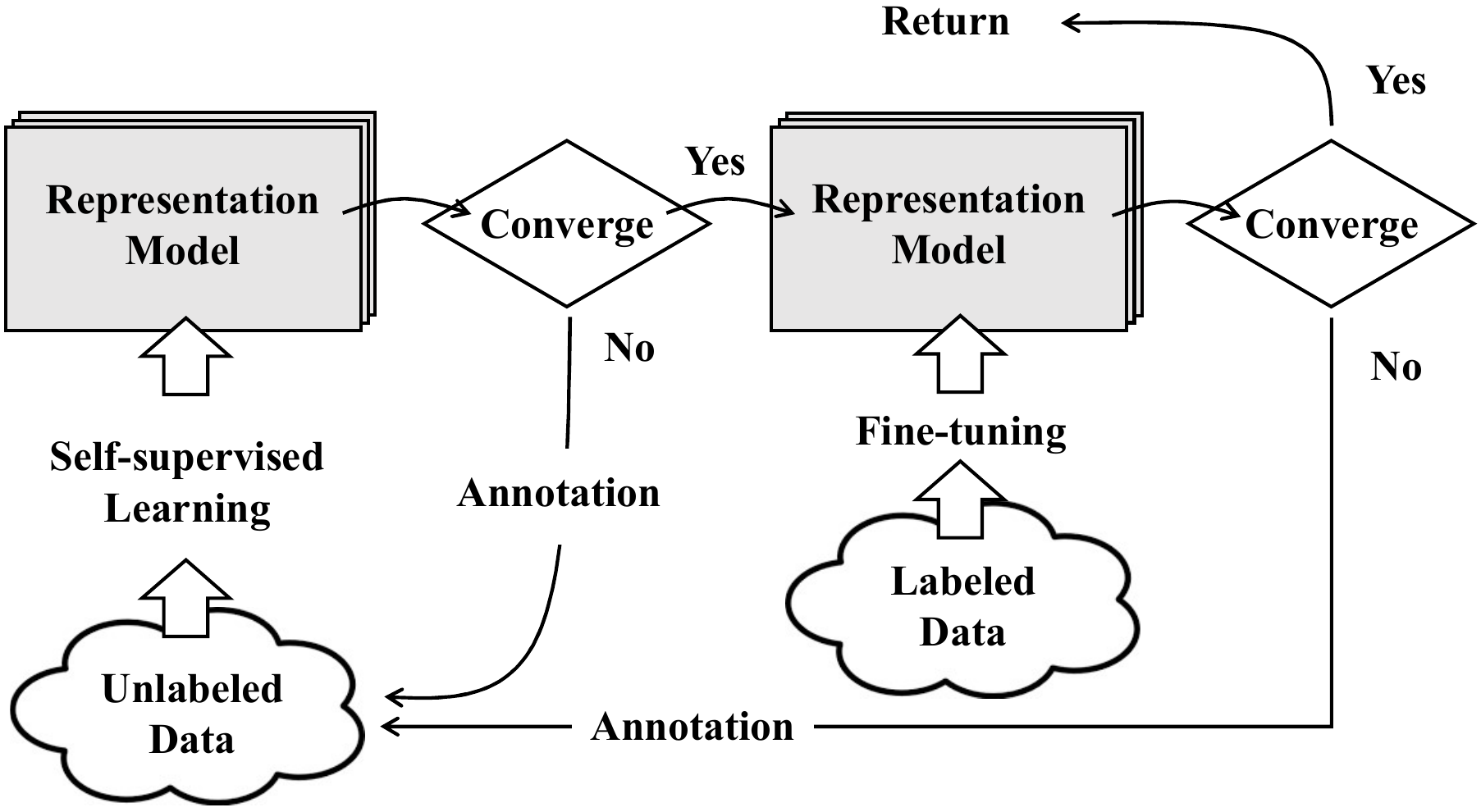}
\caption{\small InfoCSE++. Firstly, the representation model is trained by InfoCSE based on the unlabeled data. Then, the representation model is fine-tuned with the labeled data. After the fine-tuning, the representation model is sent back to perform InfoCSE once again. The iteration will terminate when the fine-tuning performance stops growing.}
\vspace{-5pt}
\label{fig:3}
\end{figure}

\subsection{InfoCSE++}
We extend the basic framework into InfoCSE++ based on the inspiration from SimCLRv2 \cite{chen2020big}, which further improves our performance in few-shot scenarios. (Figure \ref{fig:3}). 

Firstly, the representation model is trained by InfoCSE with unlabeled data. After the convergence of InfoCSE, the model will be fine-tuned on labeled data. Unlike the conventional ``\textit{one-pass pretraining and fine-tuning}'' paradigm, the fine-tuned model will be sent back and trained by InfoCSE once again; i.e., the fine-tuned model will be used as the initialization of InfoCSE (ln. 2 in Alg. \ref{alg:1}), from which a new representation model is trained based on the unlabeled data. The \textit{InfoCSE} and \textit{fine-tuning} will be iteratively carried out until the convergence. Here, we monitor the training process by checking the fine-tuned model's performance on the validation set. Once the validation performance stops growing, the best fine-tuned model will be returned as the final result. 

\renewcommand{\arraystretch}{1.25}
\newcolumntype{C}[1]{>{\centering\let\newline\\\arraybackslash\hspace{0pt}}m{#1}}
\newcommand\ChangeRT[1]{\noalign{\hrule height #1}}




\begin{table*}[t]
    \centering
    \scriptsize
    \begin{tabular}{p{1.6cm} | C{1.0cm} C{1.0cm} C{1.0cm} | C{1.0cm} C{1.0cm} C{1.0cm} | C{1.0cm} C{1.0cm} C{1.0cm} }
    \ChangeRT{1pt}
    & \multicolumn{3}{c}{\textbf{News}} &       
      \multicolumn{3}{|c}{\textbf{Web Document}} & \multicolumn{3}{|c}{\textbf{Web Browsing}} \\
    \cmidrule(lr){1-1}
    \cmidrule(lr){2-4}
    \cmidrule(lr){5-7}
    \cmidrule(lr){8-10}
    \textbf{Methods} & 
    \textbf{R@5} & \textbf{R@10} & \textbf{R@20} & \textbf{R@5} & \textbf{R@10} & \textbf{R@20} & \textbf{R@5} & \textbf{R@10} & \textbf{R@20} \\
    \hline
    SimCSE & 0.3236 & 0.3733 & 0.4744 & 0.1639 & 0.2129 & 0.2961 & 0.0319 & 0.0456 & 0.0694  \\
    ConSERT & 0.2512 & 0.2912 & 0.3839 & 0.1335 & 0.1717 & 0.2483 & 0.0295 & 0.0418 & 0.0628 \\
    Mirror-BERT & 0.2950 & 0.3399 & 0.4422 & 0.1910 & 0.2434 & 0.3333 & 0.0340 & 0.0491 & 0.0754 \\
    ICT & 0.3872 & 0.4540 & 0.5589 & 0.2842 & 0.3603 & 0.4678 & 0.0595 & 0.0855 & 0.1210  \\
    CPC & 0.3250 & 0.3819 & 0.4856 & 0.2787 & 0.3540 & 0.4608 & 0.0569 & 0.0818 & 0.1166  \\
    DeCLUTR & \underline{0.4786} & \underline{0.5380} & \underline{0.6495} & \underline{0.3155} & \underline{0.3925} & \underline{0.5104} & \underline{0.0628} & \underline{0.0900} & \underline{0.1259} \\
    CONPONO & 0.4257 & 0.5010 & 0.5980 & 0.2285 & 0.2953 & 0.3862 & 0.0142 & 0.0251 & 0.0403  \\
    \hline
    InfoCSE & \textbf{0.6345} & \textbf{0.7016} & \textbf{0.7864} & \textbf{0.3612} & \textbf{0.4425} & \textbf{0.5538} & \textbf{0.0661} & \textbf{0.0967} & \textbf{0.1362}
  \\
    \ChangeRT{1pt}
    \end{tabular}
    \vspace{-5pt}
    \caption{\small Zero-shot evaluation (our own performances are bolded, the strongest baselines are underlined). The self-supervised sentence embeddings trained on unlabeled corpora are directly used for the downstream retrieval tasks without fine-tuning.}
    \vspace{-15pt}
    \label{tab:1}
\end{table*}

\section{Experiment}

We use three \textbf{Unlabeled Corpora} for self-supervision. 1) \textit{News}, which consists of Microsoft News articles released by MIND \cite{wu2020mind}. 2) \textit{Web Document}, which contains web documents released by MS MARCO corpus \cite{nguyen2016ms}. 3) \textit{Web Browsing}, which is an industrial corpus collected from a commercial search platform: each data instance contains the sequence of a user's browsed web titles on the Internet. We also use \textbf{Labeled Corpora} for few-shot learning. For News, each news article is paired with its \textit{headline} (the unlabeled corpus only has news bodies). For Web Document, each document is paired with a \textit{search query} from Bing \cite{craswell2020overview}. For Web Browsing, each sequence is paired with an \textit{advertisement} clicked by the user. 

The performances are evaluated in terms of the \textbf{recall rates} on labeled corpora, which directly reflects the effectiveness for ad-hoc retrieval tasks. Particularly, based on sentences from the input document (news, web document, or web browsing sequence), the ground-truth counterpart (headline, search query, or ad-click) needs to be retrieved from all the candidates within the dataset. The evaluation consists of two parts: 1) \textbf{zero-shot evaluation}, where the self-supervised sentence embeddings are directly applied to the downstream retrieval task without fine-tuning (we take the average pooling of all the sentences' embeddings within the input document to derive its representation); 2) \textbf{few-shot evaluation}, where the self-supervised sentence embeddings are fine-tuned with a small amount of labeled data for the downstream retrieval task.

We make use of the latest self-supervised contrastive learning methods for evaluation, which are corresponding to the taxonomies discussed in the introduction. 1) We use \textbf{SimCSE} \cite{gao2021simcse}, \textbf{ConSERT} \cite{yan2021consert} and \textbf{Mirror-BERT} \cite{liu2021fast}, where different views of the same sentences are generated for the positive samples. 2) We use \textbf{ICT} \cite{chang2020pre,guu2020realm}, \textbf{CPC} \cite{oord2018representation} and \textbf{DeCLUTR} \cite{giorgi2020declutr}, where sentences and their contexts are used for as the positive samples. 3) We use \textbf{CONPONO} \cite{iter2020pretraining}, where neighboring sentences within the same documents are used as the positive samples. All the baseline approaches leverage BERT base \cite{Devlin2019BERT} as their text encoding backbones, which is the same as our own method for the sake of fair comparison. Supplementary results, more details about the datasets and implementations can be found in \textbf{Appendix}. Our code will be \textbf{open-sourced} soon after the review stage.

\begin{table*}[t]
    \centering
    \scriptsize
    \begin{tabular}{p{1.6cm} | C{1.0cm} C{1.0cm} C{1.0cm} | C{1.0cm} C{1.0cm} C{1.0cm} | C{1.0cm} C{1.0cm} C{1.0cm} }
    \ChangeRT{1pt}
    & \multicolumn{3}{c}{\textbf{News}} &       
      \multicolumn{3}{|c}{\textbf{Web Document}} & \multicolumn{3}{|c}{\textbf{Web Browsing}} \\
    \cmidrule(lr){1-1}
    \cmidrule(lr){2-4}
    \cmidrule(lr){5-7}
    \cmidrule(lr){8-10}
    \textbf{Methods} & 
    \textbf{1,000} & \textbf{2,000} & \textbf{5,000} & \textbf{1,000} & \textbf{2,000} & \textbf{5,000} & \textbf{1,000} & \textbf{2,000} & \textbf{5,000} \\
    \hline
    BERT & 0.3546 & 0.3742 & 0.4206 & 0.3035 & 0.3331 & 0.3717 & 0.0632 & 0.0728 & 0.0844  \\
    SimCSE & 0.4326 & 0.4460 & 0.4803 & 0.3338 & 0.3685 & 0.4036 & 0.0736 & 0.0789 & 0.0898  \\
    ConSERT & 0.4306 & 0.4604 & 0.5001 & 0.3517 & 0.3786 & 0.4225 & 0.0704 & 0.0751 & 0.0847
     \\
    Mirror-BERT & 0.4344 & 0.4508 & 0.4881 & 0.3463 & 0.3682 & 0.4085 & 0.0742 & 0.0780 & 0.0904
     \\
    ICT & 0.5351 & 0.5477 & 0.5644 & 0.4240 & 0.4358 & 0.4403 & \underline{0.0799} & 0.0830 & \underline{0.0914} \\
    CPC & 0.4349 & 0.4589 & 0.4952 & 0.4348 & 0.4513 & 0.4650 & 0.0788 & \underline{0.0844} & 0.0913  \\
    DeCLUTR & \underline{0.5824} & \underline{0.5854} & \underline{0.6080} & \underline{0.4367} & \underline{0.4525} & \underline{0.4807} & 0.0780 & 0.0832 & 0.0892  \\
    CONPONO & 0.5430 & 0.5467 & 0.5618 & 0.4000 & 0.4093 & 0.4428 & 0.0217 & 0.0233 & 0.0286  \\
    \hline
    InfoCSE & \textbf{0.6484} & \textbf{0.6611} & \textbf{0.6788} & \textbf{0.4644} & \textbf{0.4850} & \textbf{0.4957} & \textbf{0.0892} & \textbf{0.0921} & \textbf{0.0983}  \\ 
    \ChangeRT{1pt}
    \end{tabular}
    \vspace{-5pt}
    \caption{\small Few-shot evaluation (Recall@5 is reported). The self-supervised sentence embeddings are fine-tuned with small amounts of labeled data (1,000, 2,000, 5,000 positive samples) before being applied to the downstream retrieval tasks. }
    \vspace{-15pt}
    \label{tab:2}
\end{table*}




\subsection{Analysis}
The experimental studies are dedicated to the exploration of three major issues: 1) InfoCSE's impact on zero-shot sentence representation, 2) InfoCSE's impact on few-shot sentence representation with typical fine-tuning, 3) InfoCSE++'s impact on few-shot sentence representation. We also make extra studies on model initialization, clustering method, convergence analysis, and discuss our collaboration with other non-SCL pretraining of sentence representation. 

$\bullet$ \textbf{Zero-shot performance}. The experiment results are shown in Table \ref{tab:1}, where the self-supervised sentence embeddings are directly applied for the downstream retrieval tasks. According to the demonstrated results, InfoCSE consistently outperforms all the baseline methods with notable margins, which indicates its effectiveness in zero-shot scenarios. Besides, we may also observe the following interesting phenomenons. 

Firstly, although learning from different views of the same sentences (e.g., SimCSE, ConSERT, Mirror-BERT) is able to achieve reasonable zero-shot recall rates, its performances are limited in general, compared with other methods which learn from ``sentences and contexts'' (e.g., ICT) or ``different sentences'' (e.g., CONPONO). Such an observation is consistent with our previous analysis: different views of the same sentences are likely to be trivial positive samples due to high lexical overlaps, which restricts the effect of self-supervision. 

Secondly, the performances of learning from different sentences (CONPONO) are diversified on different datasets: it is a competitive baseline on News and Web Document; however, it becomes highly limited on Web Browsing. As discussed, learning from different sentences is disputed: for one thing, different sentences are lexically distinguished, which is beneficial for the learning of in-depth semantics; for another thing, different sentences are likely to be irrelevant, which will become false positive samples. As a result, its performance, in reality, all depends on the specific dataset: if sentences within each document tend to have one single or very few underlying topics, the opportunity of having false positive samples will be small, which helps to generate strong self-supervision performance; however, if sentences within each document tend to have diversified topics, the opportunity of having false positive samples will be large, which will result in severely limited performances. 

The above observations further echo the rationality about InfoCSE: it leverages different sentences from each document, which provides lexically distinguished sentence pairs; meanwhile, it is able to iteratively filter out the irrelevant sentence pairs, which are potentially false positive samples. Therefore, it may always enjoy a superior self-supervision effect in different scenarios.

\begin{table*}[t]
    \centering
    \scriptsize
    \begin{tabular}{p{2.0cm} | C{1.0cm} C{1.0cm} C{1.0cm} | C{1.0cm} C{1.0cm} C{1.0cm} | C{1.0cm} C{1.0cm} C{1.0cm} }
    \ChangeRT{1pt}
    & \multicolumn{3}{c}{\textbf{News}} &       
      \multicolumn{3}{|c}{\textbf{Web Document}} & \multicolumn{3}{|c}{\textbf{Web Browsing}} \\
    \cmidrule(lr){1-1}
    \cmidrule(lr){2-4}
    \cmidrule(lr){5-7}
    \cmidrule(lr){8-10}
    \textbf{Methods} & 
    \textbf{R@5} & \textbf{R@10} & \textbf{R@20} & \textbf{R@5} & \textbf{R@10} & \textbf{R@20} & \textbf{R@5} & \textbf{R@10} & \textbf{R@20} \\
    \hline
    Fine-tuned (1000) & 0.6484 & 0.7171 & 0.7992 & 0.4644 & 0.5419 & 0.6472 & 0.0892 & 0.1260 & 0.1727 \\
    InfoCSE++ (1000) & \textbf{0.6852} & \textbf{0.7501} & \textbf{0.8268} & \textbf{0.4853} & \textbf{0.5641} & \textbf{0.6642} & \textbf{0.0940} & \textbf{0.1308} & \textbf{0.1794} \\
    Fine-tuned (All) & 0.6927 & 0.7559 & 0.8307 & 0.5129 & 0.5979 & 0.6988 & 0.1039 & 0.1426 & 0.1936 \\
    InfoCSE++ (All) & \textbf{0.7033} & \textbf{0.7653} & \textbf{0.8407} & \textbf{0.5358} & \textbf{0.6216} & \textbf{0.7215} & \textbf{0.1041} & \textbf{0.1448} & \textbf{0.1967} \\
    \hline
    IDC (K-Means) & 0.5343 & 0.6111 & 0.7099 & 0.3366 & 0.4159 & 0.5206 & 0.0600 & 0.0875 & 0.1255 \\
    IDC (Partition) & \textbf{0.6345} & \textbf{0.7016} & \textbf{0.7864} & \textbf{0.3612} & \textbf{0.4425} & \textbf{0.5538} & \textbf{0.0661} & \textbf{0.0967} & \textbf{0.1362} \\
    \hline
    InfoCSE (Iter-1) & 0.5721 & 0.6421 & 0.7332 & 0.3531 & 0.4348 & 0.5452 & 0.0646 & 0.0939 & 0.1315 \\
    InfoCSE (Iter-2) & \textbf{0.6345} & \textbf{0.7016} & \textbf{0.7864} & \textbf{0.3612} & \textbf{0.4425} & \textbf{0.5538} & \textbf{0.0661} & \textbf{0.0967} & 0.1362 \\
    InfoCSE (Iter-3) & 0.6341 & 0.6999 & 0.7860 & 0.3458 & 0.4247 & 0.5350 & 0.0656 & 0.0956 & \textbf{0.1369} \\
    \hline
    InfoCSE (CP) & \textbf{0.6345} & \textbf{0.7016} & \textbf{0.7864} & 0.3612 & 0.4425 & \textbf{0.5538} & 0.0661 & 0.0967 & 0.1362 \\
    InfoCSE (SC) & 0.6155 & 0.6859 & 0.7742 & \textbf{0.3617} & \textbf{0.4428} & 0.5530 & \textbf{0.0733} & \textbf{0.1051} & \textbf{0.1479} \\
    InfoCSE (ICT) & 0.6302 & 0.6973 & 0.7830 & 0.3601 &  0.4420 & 0.5511 & 0.0727 & 0.1043 & 0.1454 \\
    \hline
    InfoCSE (BERT) & 0.6345 & 0.7016 & 0.7864 & 0.3612 & 0.4425 & 0.5538 & 0.0661 & 0.0967 & 0.1362 \\
    InfoCSE (SBERT) & \textbf{0.6639} & \textbf{0.7275} & \textbf{0.8070} & \textbf{0.3692} & \textbf{0.4509} & \textbf{0.5621} & \textbf{0.0764} & \textbf{0.1087} & \textbf{0.1525} \\
    \ChangeRT{1pt}
    \end{tabular}
    \vspace{-5pt}
    \caption{\small Evaluations of (1) InfoCSE++, (2) clustering methods, (3) convergence, (4) initialization, (5) integration with SBERT. The few-shot performances are reported for (1); the zero-shot performances are reported for (2-5).}
    \vspace{-15pt}
    \label{tab:3}
\end{table*}


$\bullet$ \textbf{Few-shot performance}.
The self-supervised sentence embeddings are fine-tuned with labeled data for the evaluation of few-shot performances (Table \ref{tab:2}). In our experiment, the scale of labeled data (i.e., the number of positive samples) is changed from 1000, 2000, to 5000; with more labeled data, the sentence embeddings can be better fine-tuned for the downstream retrieval tasks. Besides, we also introduce the BERT baseline for comparison, which will further reflect the benefit from the self-supervised sentence embeddings (BERT is omitted in Table \ref{tab:1} as it is incapable of making zero-shot ad-hoc retrieval).

The few-shot performances in Table \ref{tab:2} are almost consistent with the previous observations reported in Table \ref{tab:1}. Firstly, the fine-tuned InfoCSE remains the strongest approach, which outperforms all the baselines in each of the testing cases; e.g., for the fine-tuned performances with 1000 samples, InfoCSE relatively improves the Recall@5 by 
\textit{11.33}\%, \textit{6.34}\% and \textit{11.64}\% over the strongest baseline performances on different datasets. Besides, the improvements over BERT are even more remarkable, which reflects the effectiveness of InfoCSE in few-shot scenarios. Secondly, the stronger baselines in zero-shot scenarios tend to achieve higher performances in few-shot scenarios as well. Thirdly, the performance gaps are diminished when more labeled data is used. This observation is expected: the downstream retrieval task can be directly optimized through supervised learning; thus, the differences from pretraining will gradually shrink.


$\bullet$ \textbf{Learning with InfoCSE++}. The effectiveness of InfoCSE++ is evaluated by comparing with the ``Fine-tuned'' InfoCSE (the upper part in Table \ref{tab:3}). Two different settings are evaluated here: 1) only \textit{1000} labeled instances are provided; 2) \textit{All} labeled instances within the training set are used for InfoCSE++/Fine-tuning. \textit{For the first setting}, InfoCSE++ achieves substantial improvements on all the datasets. The observed improvements are easy to interpret: the representation model's accuracy is enhanced when it is fine-tuned with labeled data. Therefore, it will improve the data annotation quality, based on which more quality positive samples can be generated to enhance self-supervised learning. The enhancement of self-supervision will pave the way for further improvement of the next round of fine-tuning. 
\textit{For the second setting}, there are still notable improvements on News and Web Document, whereas the improvement on Web Browsing is not as significant. Such a distinction is probably due to the different sizes of labeled data: the labeled data is much smaller for News and Web Document (with 10,000 and 16,366 positive samples, respectively) compared with Web Browsing (with 99,217 positive samples). With limited amounts of labeled data, the representation model is probably insufficiently fine-tuned; thus, the performance can be further improved based on more effective utilization of the unlabeled data. In contrast, if the representation model is fully fine-tuned with sufficient labeled data, the additional gain from the unlabeled data will be limited. As such, we may conclude that InfoCSE++ is more of an effective learning paradigm for the few-shot scenarios, where labeled data is highly limited.

$\bullet$ \textbf{Extended Studies.} The following extended studies about InfoCSE are shown in Table \ref{tab:3}. Firstly, We analyze our partition-based \textbf{Clustering} by making a comparison with the $K$-Means based method used in DeepCluster \cite{caron2018deep} and SwaV \cite{caron2020unsupervised}: a total of $K$ ``global clusters'' are formed based on all the sentences from different documents; for each document: sentences are grouped for IDC-based on their global cluster assignments ($K$ is empirically tuned to 10). We find that IDC ($K$-means) is inferior to the default IDC with our partition-based clustering. This is mainly because the global clusters are rigid and coarse-grained, which is unfavorable to analyze the relative correlation between sentences within the same document. Note that the intra-document $K$-means is also inappropriate: it is neither efficient (which leads to unaffordable costs in our experiment), nor flexible to deal with documents with different semantic diversities. 

We analyze the \textbf{Convergence} of InfoCSE (Iter-$k$ means the self-supervision's result after the $k$-th round IDC). We find that InfoCSE is able to quickly converge within 2 rounds of IDC, which indicates that the proposed method can be efficiently trained with little additional cost. 

We evaluate different \textbf{Initialization} methods besides CONPONO (CP), including SimCSE (SC) and ICT. For News and Web Document, different initializations converge to similar zero-shot performances. For Web Browsing, the default initialization (CP) is inferior to SC and ICT. Such a difference is consistent with the observation the CP is highly limited and much worse than SC and ICT on Web Browsing (as Table \ref{tab:1}). 

As mentioned, there are many other pretrained sentence embeddings besides self-supervised contrastive learning. Our proposed method may collaborate with them for better self-supervision performances. In this place, we explore the impact of \textbf{collaborating with SBERT}\footnote{\scriptsize https://www.sbert.net/docs/pretrained\_models.html. The strongest ckpt ``all-mpnet-base-v2'' is used (pre-trained with 1 billion sentence pairs).}, where SBERT is used to provide the first-round data annotation (it was originally performed by the CONPONO initialized BERT). We find that our zero-shot performance can be further enhanced with InfoCSE (SBERT).





\section{Conclusion}
In this paper, we proposed a novel self-supervised contrastive learning framework InfoCSE, where data annotation and representation learning could be mutually reinforced to enhance the effect of self-supervision. We designed the intra-document clustering algorithm, where lexically different but semantically correlated sentences can be excavated from documents as the positive training samples. We extended the fundamental framework into InfoCSE++, which further improved the sentence embeddings' quality in few-shot scenarios. The effectiveness of our proposed methods was verified in experimental studies, where notable improvements were achieved against the SOTA self-supervised contrastive learning-based approaches. 

\newpage 

\bibliographystyle{acl_natbib}
\bibliography{anthology,acl_ref}

\clearpage

\appendix
\section{Implementation Details}
The pseudo-code of partition-based clustering is shown in Alg. \ref{alg:clustering}. First, we organize the sentences as a fully connected graph $\mathcal{G}$ by using the representation model $f(\cdot)$ according to Eq. \ref{eq:2}. The connections that don't satisfy any of the relations in Eq. \ref{eq:3} will be pruned, which will lead to graph $\mathcal{G'}$. Each pair of sentences will be included in the same cluster, if they are reachable on $\mathcal{G'}$. Finally, the resulted clusters will be returned.

\section{Dataset Specifications}
The dataset specifications are shown in Table \ref{tab:A1}. \#Document: number of documents in the unlabeled corpus. Avg. Doc. Len.: average number of sentences within each document. \#Item: number of paired counter pairs (news headline, search query, ad-click) in the dataset. {\#Train} (label), {\#Valid} (label), {\#Test} (label): number of instances within the training, validation, testing set for the labeled corpus.

\begin{algorithm}[t]
\caption{Partition-based clustering}\label{alg:clustering}
    \LinesNumbered 
    \SetKwInOut{KwIn}{Input}
    \SetKwInOut{KwOut}{Output}
    \KwIn{sentences $\mathbf{S}$, representation model $f(\cdot)$}
    \KwOut{clusters $\{\mathcal{G}_l\}_L$}
    \Begin{
        $\mathcal{G} \leftarrow$ get the fully-connected graph based on $\textbf{S}$ and $f(\cdot)$\, as Eq. \ref{eq:2}\;    
        $\mathcal{G'} \leftarrow$prune the connections in $\mathcal{G}$ unless any of the relationships in Eq. \ref{eq:3} is satisfied\;
        $\{\{\mathcal{G}_i\}\} \leftarrow \{\{s_i\}|s_i \in \mathbf{S}\}$ \;
        \For{$s_i,s_j\in\mathbf{S^2}|(i\neq j)$}{
            \If{$s_i$ is reachable from $s_j$ in $\mathcal{G'}$}
                {merge $\mathcal{G}_i$ and $\mathcal{G}_j$ into one cluster}
            
        }
        return $\{\mathcal{G}_l\}_L$\;
    }
\end{algorithm}

\begin{table}[t]
    \centering
    \scriptsize
    \begin{tabular}{p{2cm} | C{1cm}  | C{1.5cm}  | C{1cm}  }
    \ChangeRT{1pt}
    & {\textbf{News}} &  
      {\textbf{Web Document}} & {\textbf{Web Browsing}} \\
    \hline
    \#Document & 1,500,000 & 3,213,835 & 3,000,000  \\
    Avg. Doc. Len. & 19.51 & 28.90 & 43.28 \\
    \#Item & 371,940 & 372,206 & 148,333 \\
    \#Train (label) & 10,000 & 16,336 & 99,217 \\
    \#Valid (label) & 24,363 & 5,193 & 24,708  \\
    \#Test (label) & 180,000 & 203,647 & 123,551 \\
    \ChangeRT{1pt}
    \end{tabular}
    \caption{\small Specifications of datasets}
    \label{tab:A1}
\end{table}

\section{Training Details}
As shown in Table \ref{tab:training paras}, we present the hyperparameters used for InfoCSE. On all datasets, the model is pretrained for at most 0.2M epochs and finetuned for at most 20 epochs. We use an early stopping strategy on R@5 with patience of 20,000 steps for pretraining and 2 epochs for finetuning. For optimization, we use Adam (Kingma and Ba, 2014) with $\beta_1$=0.9, $\beta_2$=0.999, $\epsilon$=1e-8 and learning rate=2e-5. We set the max token length as 32 for each sentence on all datasets. To make full use of the GPU memory, we set the batch size as 3200 for pretraining and 64 for finetuning. The training is on $8\times$ Nvidia A100-40GB GPU clusters. We use Python3.6 and PyTorch 1.7.0 for implementation. The random seed of PyTorch is fixed as 42.

\begin{table}[t]
	\centering
	\scriptsize
	\begin{tabular}{cc}
		\toprule
		Optimizer& {Adam} \\ 
		Adam $\beta_1$& {0.9}  \\
		Adam $\beta_2$& {0.999} \\
		Adam $\epsilon$& {1e-8} \\
		Learning rate & 2e-5  \\
		PyTorch random seed & {42} \\
		Max pretraining steps & 0.2M \\
		Pretraining batch size & 3200 \\
		Finetuning batch size & 64 \\
        Max finetuning epochs & 20 \\ 
		\bottomrule
	\end{tabular}
	\caption{Hyper-parameters settings}
	\label{tab:training paras}
\end{table}

\section{Case Analysis}
We use the following real-world case to visualize the improvement of data annotation quality with InfoCSE (the web browsing sequence is used for demonstration).

Firstly, we have the following web browsing sequence from an online user: 

\textbf{(1)} \textit{Official LEGO® Shop US}; \textbf{(2)} \textit{LEGO® DUPLO® World People Set}; \textbf{(3)} \textit{Spencer Greece Gray Accent Chair}; \textbf{(4)} \textit{Venus Navy Accent Chair | Bobs.com}; \textbf{(5)} \textit{Calvin Onyx Black Bob-O-Pedic Queen Sleeper Sofa}; \textbf{(6)} \textit{Capri Denim Bob-O-Pedic Sleeper Sofa}; \textbf{(7)} \textit{Youth Vilano Balance Bike}; \textbf{(8)} \textit{Strider 12 Sport Balance Bike}.

While initializing the model, all neighbouring sentences are treated as positive. As a result, the following false positive samples are generated:

\textbf{(2)} \textit{LEGO® DUPLO® World People Set}; \textbf{(3)} \textit{Spencer Greece Gray Accent Chair}

\textbf{(4)} \textit{Venus Navy Accent Chair | Bobs.com}; \textbf{(5)} \textit{Calvin Onyx Black Bob-O-Pedic Queen Sleeper Sofa}

\textbf{(6)} \textit{Capri Denim Bob-O-Pedic Sleeper Sofa}; \textbf{(7)} \textit{Youth Vilano Balance Bike}\}

After initiation, the first round Intra-Document Clustering is performed, where the following sentence clusters are formed:

$\bullet$ \textbf{C1}: \{\textbf{(1)} \textit{Official LEGO® Shop US}; \textbf{(2)} \textit{LEGO® DUPLO® World People Set}\}

$\bullet$ \textbf{C2}: \{\textbf{(3)} \textit{Spencer Greece Gray Accent Chair}; \textbf{(4)} \textit{Venus Navy Accent Chair | Bobs.com}; \textbf{(5)} \textit{Calvin Onyx Black Bob-O-Pedic Queen Sleeper Sofa}; \textbf{(6)} \textit{Capri Denim Bob-O-Pedic Sleeper Sofa}\}

$\bullet$ \textbf{C3}: \{\textbf{(7)} \textit{Youth Vilano Balance Bike}; \textbf{(8)} \textit{Strider 12 Sport Balance Bike}\}

It can be observed that the previous false positives: ``\{(2), (3)\}'' and ``\{(6), (7)\}'' can be removed after the first iteration. However, ``\{(4), (5)\}'' still remains as their semantic difference is much harder to discriminate.

By taking sentences within the same cluster as positive samples, the representation model is updated. For the next iteration, the sentences are re-clustered as follows:

$\bullet$ \textbf{C1}: \{\textbf{(1)} \textit{Official LEGO® Shop US}; \textbf{(2)} \textit{LEGO® DUPLO® World People Set}\}

$\bullet$ \textbf{C2}: \{\textbf{(3)} \textit{Spencer Greece Gray Accent Chair}; \textbf{(4)} \textit{Venus Navy Accent Chair | Bobs.com}\}

$\bullet$ \textbf{C3}: \{\textbf{(5)} \textit{Calvin Onyx Black Bob-O-Pedic Queen Sleeper Sofa}; \textbf{(6)} \textit{Capri Denim Bob-O-Pedic Sleeper Sofa}\}

$\bullet$ \textbf{C4}: \{\textbf{(7)} \textit{Youth Vilano Balance Bike}; \textbf{(8)} \textit{Strider 12 Sport Balance Bike}\}

Now, all the previous false positives can be removed from the clustering result.


\end{document}